\documentclass[a4paper,twoside]{article}

\usepackage{epsfig}
\usepackage{caption}
\usepackage{subcaption}
\usepackage{calc}
\usepackage{amssymb}
\usepackage{amstext}
\usepackage{amsmath}
\usepackage{amsthm}
\usepackage{multicol}
\usepackage{multirow}
\usepackage{pslatex}
\usepackage{apalike}
\usepackage{algorithm2e}
\usepackage{float}
\usepackage[bottom]{footmisc}
\usepackage{SCITEPRESS}     % Please add other packages that you may need BEFORE the SCITEPRESS.sty package.

\begin{document}

\title{Enhancement-Driven Pretraining for Robust Fingerprint Representation Learning}

% \author{\authorname{First Author Name\sup{1}\orcidAuthor{0000-0000-0000-0000}, Second Author Name\sup{1}\orcidAuthor{0000-0000-0000-0000} and Third Author Name\sup{2}\orcidAuthor{0000-0000-0000-0000}}
% \affiliation{\sup{1}Institute of Problem Solving, XYZ University, My Street, MyTown, MyCountry}
% \affiliation{\sup{2}Department of Computing, Main University, MySecondTown, MyCountry}
% \email{\{f\_author, s\_author\}@ips.xyz.edu, t\_author@dc.mu.edu}
% }

\author{\authorname{Ekta Gavas\sup{1}\orcidAuthor{0000-0001-6437-3357}, Kaustubh Olpadkar\sup{2}\orcidAuthor{0009-0008-3811-4771}, Anoop Namboodiri\sup{1}\orcidAuthor{0000-0002-4638-0833} }
\affiliation{\sup{1}Centre for Visual Information Technology, International Institute of Information Technology, Hyderabad, India}
\affiliation{\sup{2}Stony Brook University, USA}
\email{ekta.gavas@research.iiit.ac.in, kaustubh.olpadkar@gmail.com,  anoop@iiit.ac.in}
}

\keywords{Fingerprint Representation Learning, Fingerprint Verification, Self-Supervised Learning, Deep Learning}

\abstract{Fingerprint recognition stands as a pivotal component of biometric technology, with diverse applications from identity verification to advanced search tools. In this paper, we propose a unique method for deriving robust fingerprint representations by leveraging enhancement-based pre-training. Building on the achievements of U-Net-based fingerprint enhancement, our method employs a specialized encoder to derive representations from fingerprint images in a self-supervised manner. We further refine these representations, aiming to enhance the verification capabilities. Our experimental results, tested on publicly available fingerprint datasets, reveal a marked improvement in verification performance against established self-supervised training techniques. Our findings not only highlight the effectiveness of our method but also pave the way for potential advancements. Crucially, our research indicates that it is feasible to extract meaningful fingerprint representations from degraded images without relying on enhanced samples.}

\onecolumn \maketitle \normalsize \setcounter{footnote}{0} \vfill

\section{\uppercase{Introduction}}
Fingerprint recognition remains a cornerstone in biometric identification, valued for its uniqueness, permanence, and user-friendliness \cite{Maltoni2022,Wayman2005,Allen2005}. As demand in law enforcement, personal identification, and secure authentication continues to rise, the need to enhance precision and efficiency in fingerprint recognition systems becomes increasingly vital \cite{Allen2005}.

Despite advancements in the field, challenges persist, including handling partial or distorted fingerprints, managing high interclass similarity, and addressing the expansive dimensionality of the feature space \cite{Maltoni2022,hong1998fingerprint,cappelli2007fingerprint}. Many state-of-the-art works in fingerprint matching rely on minutia-based approaches \cite{ratha1996real,chang1997fast,Maltoni2022,cappelli2010mcc,cappelli2010minutia,jain2001fingerprint,jain1997line}. This involves extracting minutiae and matching templates to determine similarity, but traditional minutia-based methods face limitations like noise sensitivity and difficulty with partial prints \cite{Maltoni2009,hong1998fingerprint,Maltoni2009,zaeri2011minutiae}.

In contrast, Convolutional Neural Networks (CNNs) present a contemporary solution, effectively overcoming limitations and improving accuracy. CNNs handle partial prints, tolerate distortions, and adapt to diverse finger conditions, showcasing scalability and efficient comparison even with growing databases \cite{nguyen2018robust,deshpande2020cnnai,darlow2017fingerprint,tang2017fingernet,engelsma2019learning}.

The surge in self-supervised learning techniques in machine learning has extended to fingerprint biometrics \cite{jaiswal2020survey,liu2021self,jing2020self}. Offering solutions to challenges in data acquisition, self-supervised learning bypasses time-consuming labeling processes. In this paper, we explore the potential of deep CNNs for superior matching performance, proposing a pretraining technique based on U-Net for fingerprint enhancement. The U-Net model, known for biomedical image segmentation \cite{ronneberger2015u}, effectively enhances fingerprints by extracting contextual information, aiming to derive compact, discriminative fingerprint embeddings. Our study pursues two objectives: proposing a pretraining technique with U-Net and assessing the efficacy of these representations through verification performance against existing self-supervised methods. We experiment with training and inference techniques to optimize the use of representations for fingerprint verification tasks. This paper aims to deepen our understanding of fingerprint recognition, inspiring future progress in this direction.

%-------------------------------------------------------------------------
\subsection{Contributions}

Here are the main contributions of this work:
\begin{enumerate}
    \item We suggest a pre-training technique with fingerprint enhancement task on our encoder and demonstrate the usefulness of this approach in representation learning in self-supervised setting.
    \item We describe a method to fine-tune the learned embeddings for fingerprint verification task.
    \item We evaluate our approach with various evaluation metrics demonstrating its effectiveness in fingerprint verification task and also provide a comparison with previous state-of-the-art self-supervised learning methods.
\end{enumerate}

\begin{figure*}[!ht]
    \centering
    \includegraphics[width=0.6\textwidth]{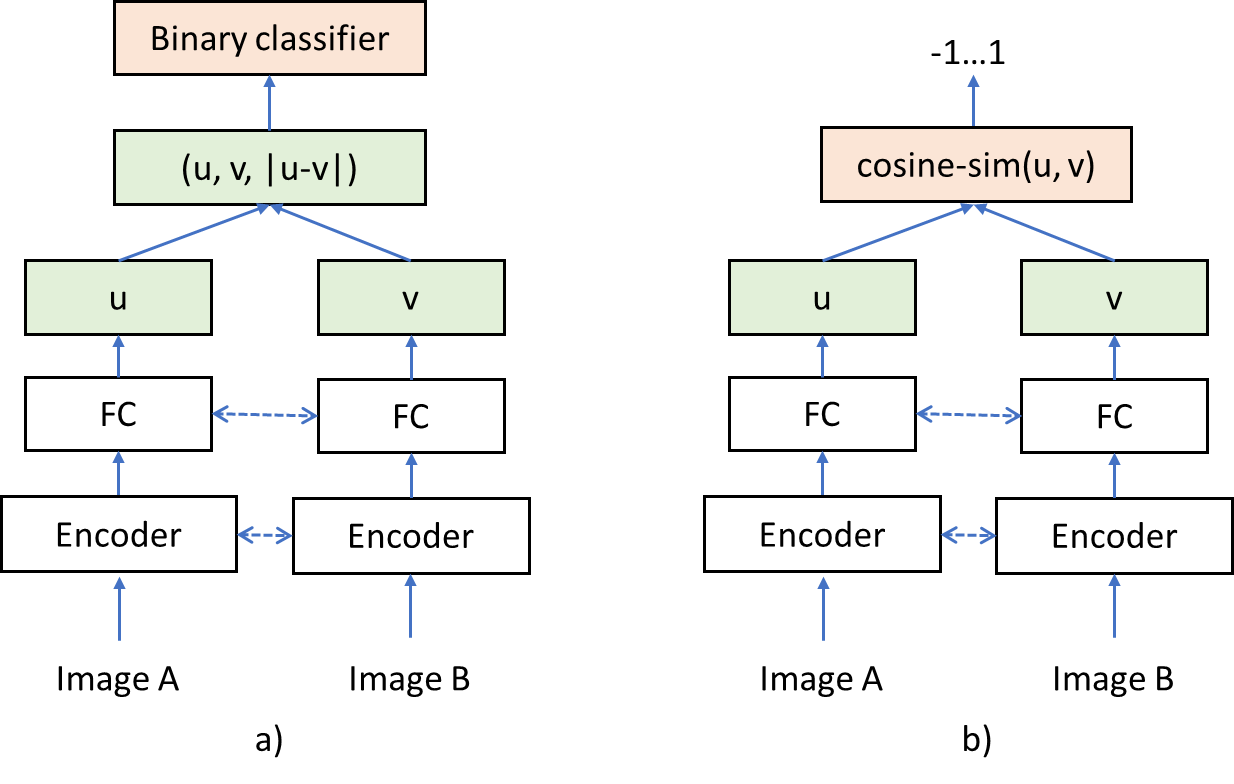}
    \caption{a) Architecture with verification objective i.e with binary classifier (at training and inference) b) Architecture to compute similarity scores (at inference). The dotted arrows indicate networks having tied weights (siamese
network structure).}
    \label{fig:arch}
\end{figure*}

\subsection{Related Work}

The need for improved fingerprint recognition tools has spurred the development of effective fingerprint representation methods. Various approaches, drawing on domain knowledge, have enhanced the accuracy and speed of fingerprint identification \cite{engelsma2019learning,tang2017fingernet}. This paper explores a pretraining technique, focusing on an enhancement task to optimize model learning for representation.

\subsubsection{Image Enhancement}

Early fingerprint image enhancement methods, such as Gabor filters and Fourier Transform, faced challenges with poor quality, noise, and pattern variations \cite{greenberg2002fingerprint,hong1998fingerprint,kim2002new,yang2002improved,liu2014latent,sherlock1992algorithm,chikkerur2005fingerprint,rahman2008improved}. Convolutional Neural Networks (CNNs), adept at hierarchical learning, have proven effective in capturing minutiae and latent features, enhancing recognition accuracy \cite{nguyen2018robust,deshpande2020cnnai,tang2017fingernet}. U-Net, originally designed for biomedical image segmentation, has been adapted for fingerprint enhancement \cite{ronneberger2015u}. Various modifications to U-Net, tailored for fingerprint enhancement tasks, have been proposed \cite{gavas2023,qian2019latent,liu2020automatic}.

\subsubsection{Self-supervised Learning Techniques}

Self-supervised learning, an alternative to traditional supervised learning, capitalizes on unlabeled data using pretext tasks for feature representation \cite{jaiswal2020survey,jing2020self}. Contrastive learning, a cornerstone of self-supervised learning, differentiates between similar and dissimilar instances \cite{liu2021self}. Techniques like SimCLR, MoCo, BYOL, SwAV, and Noise Contrastive Estimation showcase the diversity of contrastive learning approaches \cite{chen2020big,chen2020improved,grill2020bootstrap,caron2021unsupervised,gutmann2010noise}. These methods provide insights into contrastive learning's potential applications in fingerprint biometrics.

\section{\uppercase{Methodology}}
\label{sec:methodology}
\begin{figure*}
    \centering
    \includegraphics[width=0.6\textwidth]{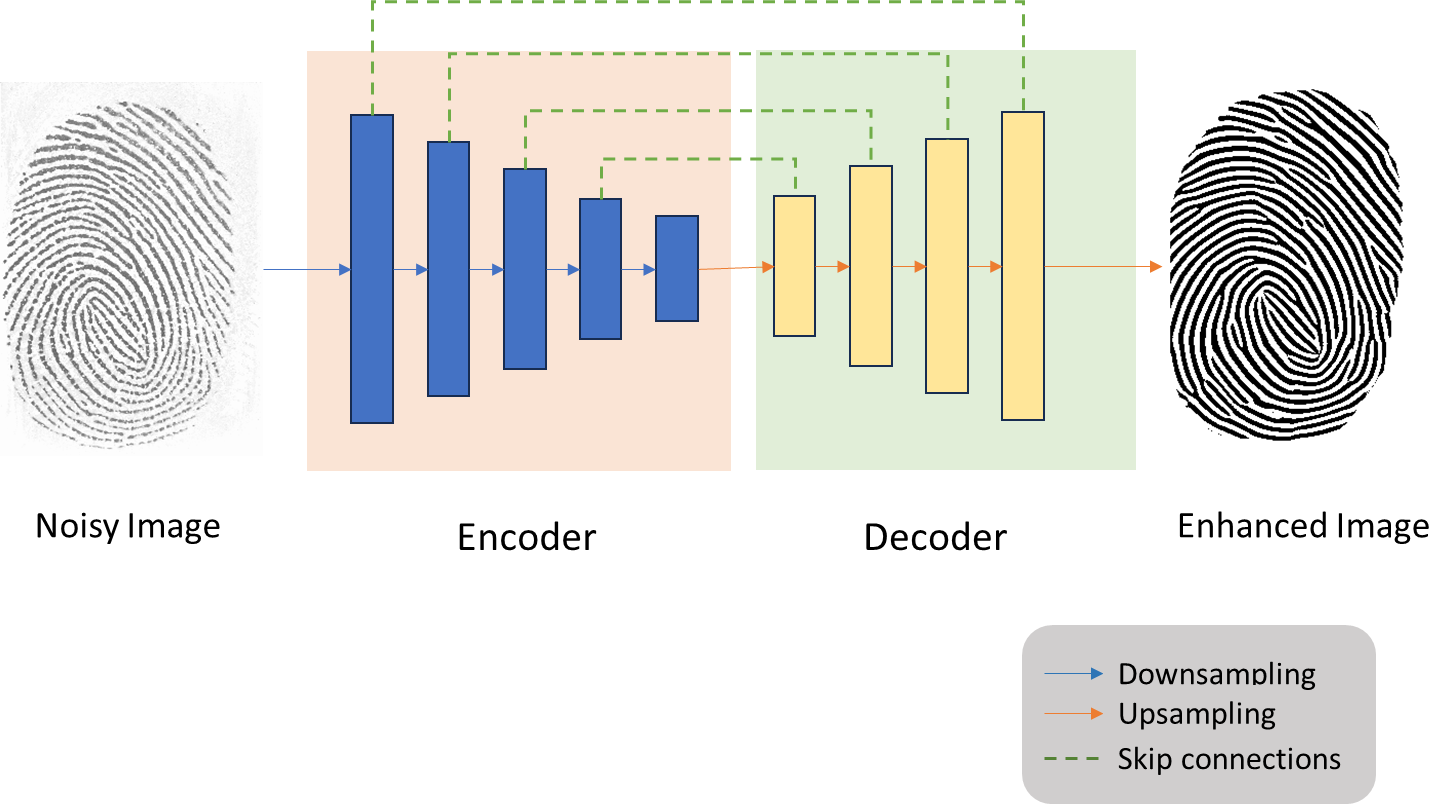}
    \caption{U-Net architecture for enhancement task for the pre-training stage in the self-supervised setting. For representation learning, the decoder is discarded and the binary classifier is attached.}
    \label{fig:unet}
\end{figure*}

The methodology for our research is constructed around a two-stage framework to probe the potential of self-supervised learning in fingerprint representation learning. A broad overview of the process is as follows:

\begin{itemize}
    \item \textbf{Stage 1: Self-Supervised Pre-training:} This is the initial stage of our methodology, in which we perform pre-training of our models in a self-supervised manner. It includes the application of both existing self-supervised learning techniques as well as our novel enhancement-based approach for this task. This stage intends to leverage the power of unlabeled data to learn meaningful representations that can serve as a starting point for subsequent stages. Notably, for all methods, we keep the encoder architecture the same. While other self-supervised methods traditionally use encoders like ResNet or Vision Transformers, in our framework we use the encoder of our U-Net-based model to ensure a fair comparison.
    
    \item \textbf{Stage 2: Probing Experiments:} Upon completion of the pre-training phase, we progress to the second stage where a few linear layers (MLP) are added on top of the frozen pre-trained encoder, making the representations 512-d. We then perform probing experiments using this newly formed model. By keeping the encoder part frozen, we ensure that the model adapts the existing representations for the verification task without altering the learned patterns from the self-supervised pre-training phase.
\end{itemize}

Following this framework, we navigate through the process of adapting and implementing self-supervised learning techniques, exploring a U-Net-based pre-training strategy, and conducting probing experiments with pre-trained networks. The sections below provide a detailed overview of the procedures involved in each stage.
% \subsection{Enhancement Encoder}
% We train the U-Net model in \ref{fig:arch} for enhancement task on 

\subsection{U-Net-based Pretraining}
While applying existing self-supervised methods to fingerprint data offers a valuable starting point, we advocate for a self-supervised learning method tailored specifically for the uniqueness of fingerprint data. Drawing on our insights from U-Net-based enhancement works, our approach employs the training of a fingerprint enhancement model as a form of self-supervision.

We employ U-Net-based fingerprint enhancement for pre-training, hypothesizing that the U-Net encoder, trained on fingerprint enhancement, holds valuable fingerprint representations. Enhancing a fingerprint image becomes an effective self-supervised task, encouraging the model to learn useful, fingerprint representations. The pre-trained encoder already encapsulates crucial information about the fingerprint, providing a foundation for further representation learning. The quality of these initial representations hinges on the efficacy of the U-Net-based enhancement model, emphasizing the significance of the model's design and training.

For enhancement-based pre-training, we use the basic U-Net architecture (Figure \ref{fig:unet}) to optimize the fingerprint enhancement task. This simple image-to-image network takes a degraded fingerprint image as input, degraded with various noises. The network aims to predict an enhanced version of the fingerprint image by removing noise while maintaining and restoring the ridge structure. This ensures the network learns minute details of fingerprint structure and enhances it where possible, aiding robust feature representation extraction later. We term it self-supervision as we use supervision from the enhancement task indirectly. This leverages a smaller amount of labeled data with limited impressions and identities.
\begin{table*}[h]
% \vspace{-5cm}
\centering
\caption{Enhancement pre-training stage results with U-Net architecture}
\label{tab:enh-results-metrics}
\centering
\begin{tabular}{|l|l|l|l|l|}
\hline
\multicolumn{1}{|c|}{\textbf{Method}} & \textbf{SSIM} & \textbf{RSME} & \textbf{PSNR} & \textbf{NFIQ2} \\ \hline
Raw Images & 0.595 & 113.23 & 6.53 & 33.42 \\ \hline
Enhancement  U-Net & 0.903 & 39.38 & 16.72 & 51.26 \\ \hline
\end{tabular}
\end{table*}

Table \ref{tab:enh-results-metrics} are the results of the first stage of our network where we are pre-training the U-Net model for enhancement task.
Results of this pre-training stage are demonstrated in Table \ref{tab:enh-results-metrics}.

\subsection{Learning Fingerprint Representation}
After the self-supervised pre-training, we conduct the probing experiments using the pre-trained networks. These experiments aim to assess the usefulness of the learned representations for the task of fingerprint verification. For this, we add a 3-layer MLP projection head on top of the frozen encoder part of the pre-trained network. We then train this model using a Sentence-BERT-like \cite{reimers2019sentencebert} siamese architecture, with a limited amount of labeled data for the fingerprint verification task. We concatenate the fingerprint representations \(u\) and \(v\) of the image pair with the element-wise difference \(|{u - v}|\) and then pass it through the linear layers and train it for binary-classification objective as illustrated in Figure \ref{fig:arch}. By keeping the encoder part frozen, the model learns to adapt the existing representations for the verification task, without changing the underlying learned patterns. This approach allows us to leverage a large amount of unlabeled data to learn initial representations and a limited amount of labeled data for supervised adaptation. Note that in the supervised fine-tuning, allowing modifications in the encoder weights can lead to higher performance on the end task, which is the future scope of this work. As our goal here is to examine the robustness of the learned representations by different pre-training techniques, we keep the encoder frozen. In summary, the combination of self-supervised pre-training with supervised fine-tuning offers a promising learning framework for fingerprint biometrics. Our methodology aims to leverage the strengths of both self-supervised and supervised learning, offering a pathway towards robust, efficient, and data-savvy fingerprint biometrics systems.

\section{\uppercase{Experiments}}

In this section, we discuss the experiments performed to evaluate our proposed approach's efficacy. We cover the specifics of our experimental setup, including the datasets used, the training details, and the evaluation metrics employed.

\subsection{Datasets and Preprocessing}

% The datasets used in this study consist of both synthetic and real-world fingerprint images, originating from the Synthetic Fingerprint Generator (SFinGe) \cite{Cappelli2004SFinGeA}, the Fingerprint Verification Competition (FVC) \cite{maiofvc2000,maio2002fvc2002,10.1007/978-3-540-25948-0_1}, and the NIST SD-302 database \cite{sd302}.

% \begin{table*}[h]
% \caption{Summary of Datasets}
% \label{table:1}
% \centering
% \begin{tabular}{|c|c|c|c|c|}
% \hline
% \textbf{Dataset} & \textbf{Source} & \textbf{Identities} & \textbf{Images} & \textbf{Purpose} \\
% \hline
% SFinGe & Synthetic & 3,700 & 15000 & Train \\
% \hline
% FVC-2000 & Real-world & 440 & 3520 & Train \\
% \hline
% FVC-2002 & Real-world & 440 & 3520 & Train \\
% \hline
% NIST SD-302 & Real-world & 2000 & 8000 & Train \\
% \hline
% SFinGe & Synthetic & 1584 & 6336 & Test \\
% \hline
% FVC-2004 & Real-world & 440 & 3520 & Test \\
% % \hline
% % NIST SD-302 & Real-world & 200 & 800 & Testing \\
% \hline
% \end{tabular}
% \end{table*}

\begin{figure*}
    \centering
    \includegraphics[width=0.6\textwidth]{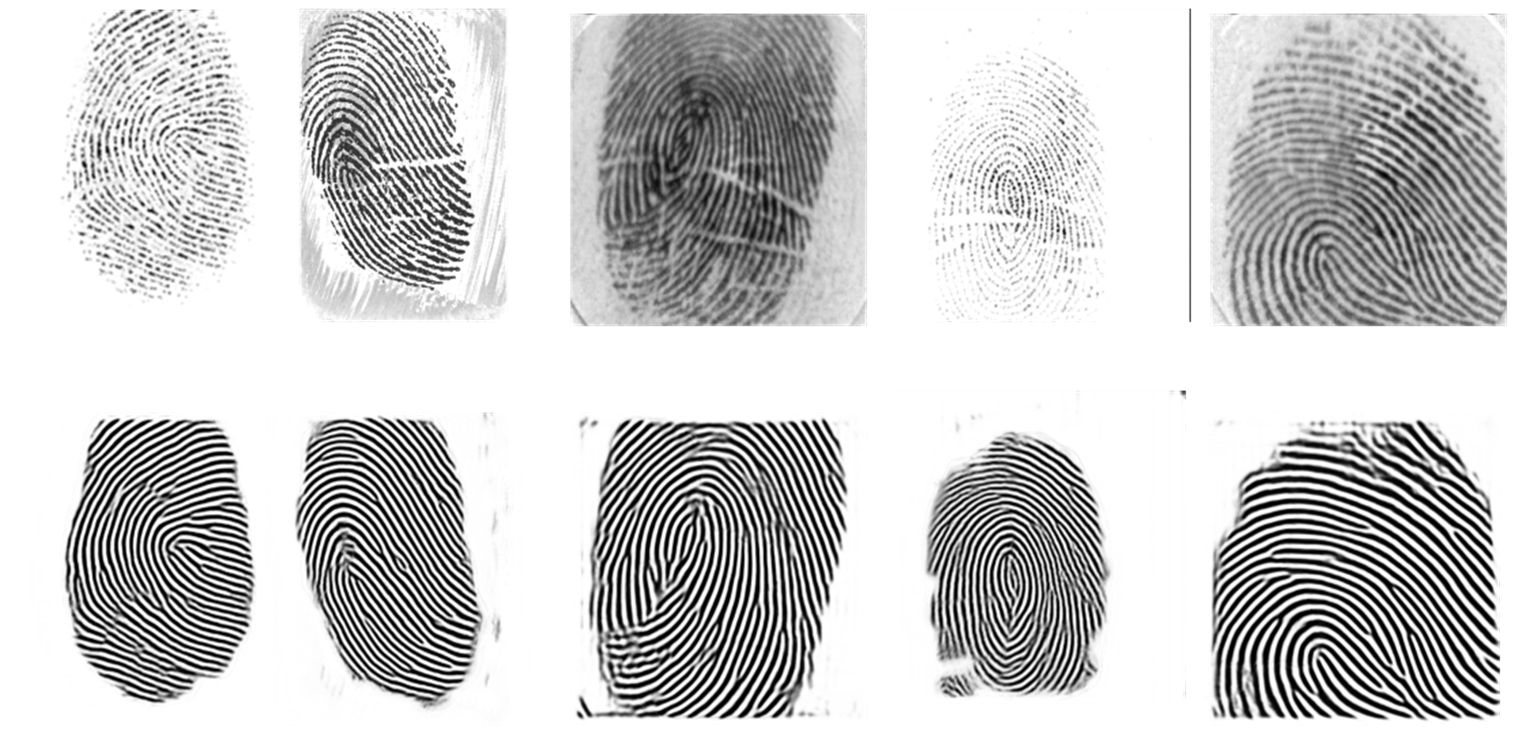}
    \caption{Degraded (top row) and Enhanced (bottom row) image pairs on FVC dataset from enhancement pre-training}
    \label{fig:enh_res}
\end{figure*}

\begin{table*}[!h]
\vspace{0.5cm}
\caption{Verification accuracy on SFinGe test dataset with genuine and imposter pairs}
\label{tab:sfinge-acc}
\centering
% \resizebox{\textwidth}{!}{%
\begin{tabular}{|crrrrrr|}
\hline
\multicolumn{7}{|c|}{\textbf{SFinGe - Accuracy}} \\ \hline
\multicolumn{1}{|c|}{\multirow{2}{*}{\textbf{Method}}} &
  \multicolumn{3}{c|}{\textbf{Classification}} &
  \multicolumn{3}{c|}{\textbf{Similarity}} \\ \cline{2-7} 
\multicolumn{1}{|c|}{} &
  \multicolumn{1}{c|}{\textbf{Imposter}} &
  \multicolumn{1}{c|}{\textbf{Genuine}} &
  \multicolumn{1}{c|}{\textbf{Entire Data}} &
  \multicolumn{1}{c|}{\textbf{Imposter}} &
  \multicolumn{1}{c|}{\textbf{Genuine}} &
  \multicolumn{1}{c|}{\textbf{Entire Data}} \\ \hline
\multicolumn{1}{|c|}{\textbf{SimCLR}} &
  \multicolumn{1}{r|}{0.968} &
  \multicolumn{1}{r|}{0.881} &
  \multicolumn{1}{r|}{0.946} &
  \multicolumn{1}{r|}{0.982} &
  \multicolumn{1}{r|}{0.749} &
  0.923 \\ \hline
\multicolumn{1}{|c|}{\textbf{SimSiam}} &
  \multicolumn{1}{r|}{0.972} &
  \multicolumn{1}{r|}{0.362} &
  \multicolumn{1}{r|}{0.819} &
  \multicolumn{1}{r|}{0.888} &
  \multicolumn{1}{r|}{0.648} &
  0.828 \\ \hline
\multicolumn{1}{|c|}{\textbf{MoCo}} &
  \multicolumn{1}{r|}{0.963} &
  \multicolumn{1}{r|}{0.881} &
  \multicolumn{1}{r|}{0.942} &
  \multicolumn{1}{r|}{0.955} &
  \multicolumn{1}{r|}{0.845} &
  0.927 \\ \hline
\multicolumn{1}{|c|}{\textbf{BYOL}} &
  \multicolumn{1}{r|}{0.96} &
  \multicolumn{1}{r|}{0.825} &
  \multicolumn{1}{r|}{0.926} &
  \multicolumn{1}{r|}{0.963} &
  \multicolumn{1}{r|}{0.718} &
  0.901 \\ \hline
\multicolumn{1}{|c|}{\textbf{Ours}} &
  \multicolumn{1}{r|}{\textbf{0.982}} &
  \multicolumn{1}{r|}{\textbf{0.886}} &
  \multicolumn{1}{r|}{\textbf{0.958}} &
  \multicolumn{1}{r|}{\textbf{0.975}} &
  \multicolumn{1}{r|}{\textbf{0.847}} &
  \textbf{0.943} \\ \hline
\end{tabular}%
% }
\end{table*}

\begin{table*}[!h]
% \vspace{0.5cm}
\caption{F1 score on SFinGe test dataset with genuine and imposter pairs}
\label{tab:sfinge-f1}
\centering
% \resizebox{\textwidth}{!}{%
\begin{tabular}{|crrrrrr|}
\hline
\multicolumn{7}{|c|}{\textbf{SFinGe - F1 score}} \\ 
\hline
\multicolumn{1}{|c|}{\multirow{2}{*}{\textbf{Method}}} & \multicolumn{3}{c|}{\textbf{Classification}} & \multicolumn{3}{c|}{\textbf{Similarity}} \\ \cline{2-7} 
\multicolumn{1}{|c|}{} &
  \multicolumn{1}{c|}{\textbf{Imposter}} &
  \multicolumn{1}{c|}{\textbf{Genuine}} &
  \multicolumn{1}{c|}{\textbf{Entire Data}} &
  \multicolumn{1}{c|}{\textbf{Imposter}} &
  \multicolumn{1}{c|}{\textbf{Genuine}} &
  \multicolumn{1}{c|}{\textbf{Entire Data}} \\ \hline
\multicolumn{1}{|c|}{\textbf{SimCLR}} &
  \multicolumn{1}{r|}{0.98} &
  \multicolumn{1}{r|}{0.8} &
  \multicolumn{1}{r|}{0.803} &
  \multicolumn{1}{r|}{0.98} &
  \multicolumn{1}{r|}{0.78} &
  0.781 \\ \hline
\multicolumn{1}{|c|}{\textbf{SimSiam}} &
  \multicolumn{1}{r|}{0.96} &
  \multicolumn{1}{r|}{0.44} &
  \multicolumn{1}{r|}{0.442} &
  \multicolumn{1}{r|}{0.92} &
  \multicolumn{1}{r|}{0.47} &
  0.469 \\ \hline
\multicolumn{1}{|c|}{\textbf{MoCo}} &
  \multicolumn{1}{r|}{0.98} &
  \multicolumn{1}{r|}{0.79} &
  \multicolumn{1}{r|}{0.785} &
  \multicolumn{1}{r|}{0.97} &
  \multicolumn{1}{r|}{0.74} &
  0.737 \\ \hline
\multicolumn{1}{|c|}{\textbf{BYOL}} &
  \multicolumn{1}{r|}{0.97} &
  \multicolumn{1}{r|}{0.74} &
  \multicolumn{1}{r|}{0.742} &
  \multicolumn{1}{r|}{0.97} &
  \multicolumn{1}{r|}{0.69} &
  0.689 \\ \hline
\multicolumn{1}{|c|}{\textbf{Ours}} &
  \multicolumn{1}{r|}{\textbf{0.99}} &
  \multicolumn{1}{r|}{\textbf{0.86}} &
  \multicolumn{1}{r|}{\textbf{0.858}} &
  \multicolumn{1}{r|}{\textbf{0.98}} &
  \multicolumn{1}{r|}{\textbf{0.81}} &
  \textbf{0.821} \\ \hline
\end{tabular}%
% }
\end{table*}

\begin{table*}
\vspace{0.5cm}
\caption{Verification accuracy on FVC test dataset with genuine and imposter pairs}
\label{tab:fvc-acc}
\centering
% \resizebox{\textwidth}{!}{%
\begin{tabular}{|crrrrrr|}
\hline
\multicolumn{7}{|c|}{\textbf{FVC - Accuracy}} \\ \hline
\multicolumn{1}{|c|}{\multirow{2}{*}{\textbf{Method}}} &
  \multicolumn{3}{c|}{\textbf{Classification}} &
  \multicolumn{3}{c|}{\textbf{Similarity}} \\ \cline{2-7} 
\multicolumn{1}{|c|}{} &
  \multicolumn{1}{c|}{\textbf{Imposter}} &
  \multicolumn{1}{c|}{\textbf{Genuine}} &
  \multicolumn{1}{c|}{\textbf{Entire Data}} &
  \multicolumn{1}{c|}{\textbf{Imposter}} &
  \multicolumn{1}{c|}{\textbf{Genuine}} &
  \multicolumn{1}{c|}{\textbf{Entire Data}} \\ \hline
\multicolumn{1}{|c|}{\textbf{SimCLR}} &
  \multicolumn{1}{r|}{0.915} &
  \multicolumn{1}{r|}{0.619} &
  \multicolumn{1}{r|}{0.841} &
  \multicolumn{1}{r|}{0.943} &
  \multicolumn{1}{r|}{0.537} &
  0.841 \\ \hline
\multicolumn{1}{|c|}{\textbf{SimSiam}} &
  \multicolumn{1}{r|}{0.956} &
  \multicolumn{1}{r|}{0.122} &
  \multicolumn{1}{r|}{0.747} &
  \multicolumn{1}{r|}{0.387} &
  \multicolumn{1}{r|}{0.733} &
  0.473 \\ \hline
\multicolumn{1}{|c|}{\textbf{MoCo}} &
  \multicolumn{1}{r|}{0.902} &
  \multicolumn{1}{r|}{0.522} &
  \multicolumn{1}{r|}{0.807} &
  \multicolumn{1}{r|}{0.896} &
  \multicolumn{1}{r|}{0.56} &
  0.812 \\ \hline
\multicolumn{1}{|c|}{\textbf{BYOL}} &
  \multicolumn{1}{r|}{0.886} &
  \multicolumn{1}{r|}{0.568} &
  \multicolumn{1}{r|}{0.806} &
  \multicolumn{1}{r|}{0.926} &
  \multicolumn{1}{r|}{0.477} &
  0.813 \\ \hline
\multicolumn{1}{|c|}{\textbf{Ours}} &
  \multicolumn{1}{r|}{\textbf{0.957}} &
  \multicolumn{1}{r|}{\textbf{0.73}} &
  \multicolumn{1}{r|}{\textbf{0.900}} &
  \multicolumn{1}{r|}{\textbf{0.933}} &
  \multicolumn{1}{r|}{\textbf{0.818}} &
  \textbf{0.904} \\ \hline
\end{tabular}%
% }
\end{table*}

\begin{table*}[!h]
\vspace{0.5cm}
\caption{F1 score on FVC test dataset with genuine and imposter pairs}
\label{tab:fvc-f1}
\centering
% \resizebox{\textwidth}{!}{%
\begin{tabular}{|crrrrrr|}
\hline
\multicolumn{7}{|c|}{\textbf{FVC - F1 score}} \\ \hline
\multicolumn{1}{|c|}{\multirow{2}{*}{\textbf{Method}}} &
  \multicolumn{3}{c|}{\textbf{Classification}} &
  \multicolumn{3}{c|}{\textbf{Similarity}} \\ \cline{2-7} 
\multicolumn{1}{|c|}{} &
  \multicolumn{1}{c|}{\textbf{Imposter}} &
  \multicolumn{1}{c|}{\textbf{Genuine}} &
  \multicolumn{1}{c|}{\textbf{Entire Data}} &
  \multicolumn{1}{c|}{\textbf{Imposter}} &
  \multicolumn{1}{c|}{\textbf{Genuine}} &
  \multicolumn{1}{c|}{\textbf{Entire Data}} \\ \hline
\multicolumn{1}{|c|}{\textbf{SimCLR}} &
  \multicolumn{1}{r|}{0.94} &
  \multicolumn{1}{r|}{0.5} &
  \multicolumn{1}{r|}{0.502} &
  \multicolumn{1}{r|}{0.95} &
  \multicolumn{1}{r|}{0.51} &
  0.51 \\ \hline
\multicolumn{1}{|c|}{\textbf{SimSiam}} &
  \multicolumn{1}{r|}{0.94} &
  \multicolumn{1}{r|}{0.16} &
  \multicolumn{1}{r|}{0.156} &
  \multicolumn{1}{r|}{0.55} &
  \multicolumn{1}{r|}{0.19} &
  0.186 \\ \hline
\multicolumn{1}{|c|}{\textbf{MoCo}} &
  \multicolumn{1}{r|}{0.93} &
  \multicolumn{1}{r|}{0.42} &
  \multicolumn{1}{r|}{0.417} &
  \multicolumn{1}{r|}{0.92} &
  \multicolumn{1}{r|}{0.43} &
  0.431 \\ \hline
\multicolumn{1}{|c|}{\textbf{BYOL}} &
  \multicolumn{1}{r|}{0.92} &
  \multicolumn{1}{r|}{0.42} &
  \multicolumn{1}{r|}{0.421} &
  \multicolumn{1}{r|}{0.94} &
  \multicolumn{1}{r|}{0.43} &
  0.432 \\ \hline
\multicolumn{1}{|c|}{\textbf{Ours}} &
  \multicolumn{1}{r|}{\textbf{0.97}} &
  \multicolumn{1}{r|}{\textbf{0.68}} &
  \multicolumn{1}{r|}{\textbf{0.679}} &
  \multicolumn{1}{r|}{\textbf{0.96}} &
  \multicolumn{1}{r|}{\textbf{0.66}} &
  \textbf{0.659} \\ \hline
\end{tabular}%
% }
\end{table*}

This study employs datasets comprising synthetic and real-world fingerprint images from SFinGe \cite{Cappelli2004SFinGeA}, FVC \cite{maiofvc2000,maio2002fvc2002,10.1007/978-3-540-25948-0_1}, and NIST SD-302 \cite{sd302}. SFinGe simulates real-world challenges, while FVC and NIST SD-302 offer large-scale, realistic fingerprint data for generalizability. This dataset combination enables model training and evaluation under diverse conditions. Synthetic data provides scalability and control, while real-world data ensures applicability. During self-supervised pre-training, only training dataset fingerprint images are used, without labels. Ground truth is needed for the enhancement task, obtained from clean images for SFinGe and generated for NIST SD-302 and FVC using a classical approach \cite{hong1998fingerprint}. The next phase involves a binary classification task for fingerprint verification. Data augmentation, vital for self-supervised learning, employs random transformations like rotation, color jitter, resize, crop, and Gaussian blur.

\subsection{Implementation Details}
We perform experiments using the PyTorch \cite{paszke2017automatic} framework on an Nvidia GeForce RTX 2080 Ti GPU for training.

Our proposed enhancement-based pre-training utilizes the U-Net architecture. This U-Net encoder is employed consistently for pre-training with other self-supervised methods to ensure fair comparison. The U-Net has a depth of 5 layers, each with 2 convolutions, and expects 512 x 512-pixel grayscale fingerprint images. The encoder outputs a 4096-dimensional vector bottleneck, reduced to 512-d with an MLP projection head. Depth-wise convolutions minimize parameters. We use $L_2$ loss for U-Net's enhancement-based pre-training, adopting losses described in respective papers for other techniques.

For pre-training with existing self-supervised methods, a grid search identifies optimal hyperparameters. Models are pre-trained for 50 epochs with early stopping.

In probing experiments, MLP projection head weights are adapted for verification while keeping encoder weights fixed. We create 1:3 positive-to-negative pairs for training and testing verification sets from each dataset. After training, models are evaluated on test sets, reporting metrics like verification accuracy, precision, recall, and F1-score. Results are presented in two ways: 1) using the binary classifier over the MLP projection head (Figure \ref{fig:arch}-a) and 2) utilizing representations with thresholds on cosine similarity (Figure \ref{fig:arch}-b). The first method evaluates the model as an end-to-end verification network, while the second explores the potential of learned representations for similarity search and recognition tasks.

\begin{figure*}[t]
\centering
\begin{subfigure}{.5\textwidth}
  \centering
  \includegraphics[width=0.9\linewidth]{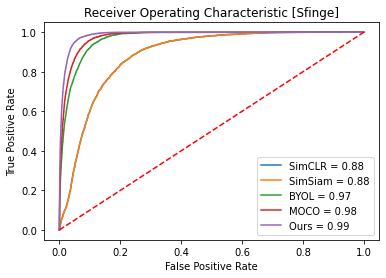}
  \label{fig:sub1}
\end{subfigure}%
\begin{subfigure}{.5\textwidth}
  \centering
  \includegraphics[width=0.9\linewidth]{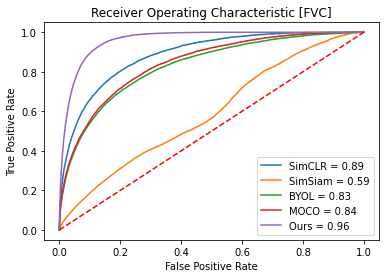}
  % \caption{A subfigure}
  \label{fig:sub2}
\end{subfigure}
\caption{ROC curve based on similarity scores on SFinGe dataset(left) and FVC dataset (right)}
\label{fig:roc-curve-sfinge}
\end{figure*}

% \subsection{Datasets}
% \subsection{Training}
\subsection{Results}
The models are first pre-trained to learn fingerprint representations using the enhancement-based approach and various self-supervised learning strategies. Because these representations are not explicitly trained for fingerprint verification or identification, using them directly for evaluation is inappropriate. To gauge the stability and usefulness of these learned representations, we add linear layers to the frozen pre-trained encoders and then train the models for fingerprint verification tasks. The encoders remain frozen, allowing only the weights of the MLP to adjust to the task, keeping the original representations unchanged. This setup aids in comparing the efficacy of different self-supervised learning techniques against our method. The results of our probing experiments are presented in Table \ref{tab:fvc-acc} (Verification Accuracy) and \ref{tab:fvc-f1} (F1-score). The verification accuracy and F1-score on the SFinGe test set are shown in Tables \ref{tab:sfinge-acc} and \ref{tab:sfinge-f1} respectively. 
% Precision and Recall results are shown in Appendix \ref{apx: sfinge_res}. 
Figure \ref{fig:enh_res} shows a few sample pairs of input and predicted images from the pre-trained U-Net model on the enhancement task used in our approach.

Our approach is compared with methods like SimCLR v2, SimSiam, MoCo v2, and BYOL on the SFinGe and FVC test sets for fingerprint verification. Verification accuracy serves as the evaluation metric for each method. The test data for fingerprint verification consists of a 1:3 ratio of positive to negative pairs, setting the random guess accuracy at 75\%. Verification accuracy is measured in two ways as described before. This is presented in the below tables under the `Classifier' column. The second way is represented under the `Similarity' column in the tables. Moreover, we also report the ROC curves in Figure \ref{fig:roc-curve-sfinge} for both datasets.

As seen from the results, our enhancement-based pre-training method consistently outperforms other self-supervised strategies across both test sets. SimCLRv2 also consistently performs well. SimSiam and BYOL methods show comparatively poor performance. It is noteworthy that all models perform better on the SFinGe test set than on the FVC test set. We believe this is due to two primary factors: the training sets contain more data from SFinGe than FVC, potentially resulting in a bias towards the former, and SFinGe is a synthetic dataset while FVC consists of real fingerprints, making the latter more challenging. Hence, the performance of models on FVC data is the real measure of the efficacy of models. Importantly, our method also provides superior performance when verification is based on the similarity of the representations, suggesting that the learned representations are also useful for fingerprint recognition.

\section{\uppercase{Limitations and Future Work}}
Despite promising results, our model demonstrates greater efficacy on the synthetic SFinGe dataset than on the real-world FVC dataset. This could be attributed to potential bias from underrepresentation of FVC data in training sets and complexities in real-life fingerprint data. Another limitation is the lack of specific training and evaluation for the fingerprint recognition task. While our model shows potential, a dedicated evaluation is essential for a comprehensive understanding of its performance. The effectiveness of self-supervised learning relies on data quality and diversity, and our study used linear probing, leaving room to explore alternative approaches like softmax or ArcFace-based classification.

Future work should address limitations by incorporating a more diverse set of real-world fingerprint datasets during training. Exploring the option of training the encoder with a smaller learning rate, rather than freezing it, could enhance generalizability and robustness. Specific training and evaluation for the recognition task, investigating alternative linear probing techniques, and exploring various self-supervised learning methods are valuable directions for further optimization.

\section{\uppercase{Conclusion}}
In this study, we explored diverse self-supervised learning techniques to pre-train a model for effective fingerprint representations in recognition and verification. A novel approach involved leveraging fingerprint enhancement as a self-supervised pre-training method. Probing experiments assessed the effectiveness of learned representations across various pre-training strategies. Comparisons against SimCLR v2, SimSiam, MoCo v2, and BYOL methods on SFinGe and FVC datasets consistently showed our method's superior verification performance. Notably, our model excelled in similarity-based verification, underscoring its effectiveness in fingerprint recognition tasks. However, models performed better on the synthetic SFinGe dataset, hinting at potential bias in the training set and real-world data complexities. Future work will expand to diverse real-world fingerprint datasets, improving model generalizability. We'll also explore additional self-supervised methods for enhanced adaptability to real-world complexities, emphasizing the potential of self-supervised learning in fingerprint biometrics while pointing to areas for exploration and refinement.

% \section*{\uppercase{Acknowledgements}}

% If any, should be placed before the references section
% without numbering. To do so please use the following command:
% \textit{$\backslash$section*\{ACKNOWLEDGEMENTS\}}

\bibliographystyle{apalike}
{\small
\bibliography{example}}

\end{document}